%% file: root.tex
\documentclass[letterpaper, 10 pt, conference]{ieeeconf}  % Comment this line out if you need a4paper

\IEEEoverridecommandlockouts                              % This command is only needed if 
\usepackage{amsmath} % assumes amsmath package installed
\usepackage{amssymb}  % assumes amsmath package installed
\usepackage{color}

\usepackage{graphicx}
\usepackage{bbold}
\usepackage{multirow}
\usepackage{subcaption}
\usepackage{tabularx}

\newcommand{\bE}{\begin{eqnarray*}}
  \newcommand{\eE}{\end{eqnarray*}}
  \newcommand{\bEn}{\begin{eqnarray}}
  \newcommand{\eEn}{\end{eqnarray}}

\newcommand{\prob}[1]{\text{p}\left(#1\right)}
\newcommand{\conprob}[2]{\prob{#1|#2}}

\title{\LARGE \bf
Estimation of Confidence Bounds in Binary Classification \\using Wilson Score Kernel Density Estimation
}

\author{Thorbjørn Mosekjær Iversen, Zebin Duan, and Frederik Hagelskjær
\thanks{All authers are with SDU Robotics, The Maersk Mc-Kinney Moller Institute, University of Southern {\tt\small \{thmi,zeb,frhag\}@mmmi.sdu.dk}}%
}% <-this % stops a space

\begin{document}

\maketitle
\thispagestyle{empty}
\pagestyle{empty}

%%%%%%%%%%%%%%%%%%%%%%%%%%%%%%%%%%%%%%%%%%%%%%%%%%%%%%%%%%%%%%%%%%%%%%%%%%%%%%%%

\input{abstract.tex}
\input{introduction.tex}
\input{relatedwork.tex}
\input{method.tex}
\input{experiments.tex}

\input{conclusion.tex}


%%%%%%%%%%%%%%%%%%%%%%%%%%%%%%%%%%%%%%%%%%%%%%%%%%%%%%%%%%%%%%%%%%%%%%%%%%%%%%%%

%%%%%%%%%%%%%%%%%%%%%%%%%%%%%%%%%%%%%%%%%%%%%%%%%%%%%%%%%%%%%%%%%%%%%%%%%%%%%%%%



% \section*{ACKNOWLEDGMENT}
% We should discuss this. It is at least NOVO and MADE, but unclear how happy either of them are to have the other included.

% {\small
\bibliographystyle{IEEEtran}
\bibliography{references}
% }

\end{document}

%% file: abstract.tex
\begin{abstract}

The performance and ease of use of deep learning-based binary classifiers have improved significantly in recent years. This has opened up the potential for automating critical inspection tasks, which have traditionally only been trusted to be done manually. However, the application of binary classifiers in critical operations depends on the estimation of reliable confidence bounds such that system performance can be ensured up to a given statistical significance. We present Wilson Score Kernel Density Classification, which is a novel kernel-based method for estimating confidence bounds in binary classification. The core of our method is the Wilson Score Kernel Density Estimator, which is a function estimator for estimating confidence bounds in Binomial experiments with conditionally varying success probabilities. Our method is evaluated in the context of selective classification on four different datasets, illustrating its use as a classification head of any feature extractor, including vision foundation models. Our proposed method shows similar performance to Gaussian Process Classification, but at a lower computational complexity.

\end{abstract}

%% file: introduction.tex
\section{INTRODUCTION}

When a robotic operation requires a high level of confidence in the success of an operation, it is common to include a verification step in the process. The purpose of this verification step is to classify the process as either successful or failed, based on some input signal, such as force feedback or images from a vision sensor. This verification is, in many cases, straightforward to implement using modern deep learning models.

There are, however, still challenges in applying modern deep learning models in critical applications, which, either due to safety or economic risks, have strong restrictions on how often it is allowed to fail. While a deep learning model might easily solve the classification task and provide an accompanying estimate of its own confidence, it is well known that these confidence estimates tend to be too optimistic, leading to untrustworthy classification. While there exist methods for calibrating these confidence scores, there is still an unmet need for classification methods that provide statistically sound confidence bounds on the classification estimates.

In this work, we propose a novel kernel-based binary classification method which, under the usual assumption of smoothness of the feature space, provides statistically sound confidence bounds on the per-instance probability of the predicted class. Our method is based on the formulation of binary classification as function estimation, where the estimated function is the probability of a positive outcome conditioned on the feature to be classified. Using this formulation, the upper and lower bounds of the probability function can be estimated using Wilson Score Kernel Density Estimation (WS-KDE), which is a frequentist method that combines kernel smoothing with the Wilson Score method for estimating confidence bounds on the $p$-value in Bernoulli trials.

\begin{figure}[t]
    \centering
    \includegraphics[width=\linewidth]{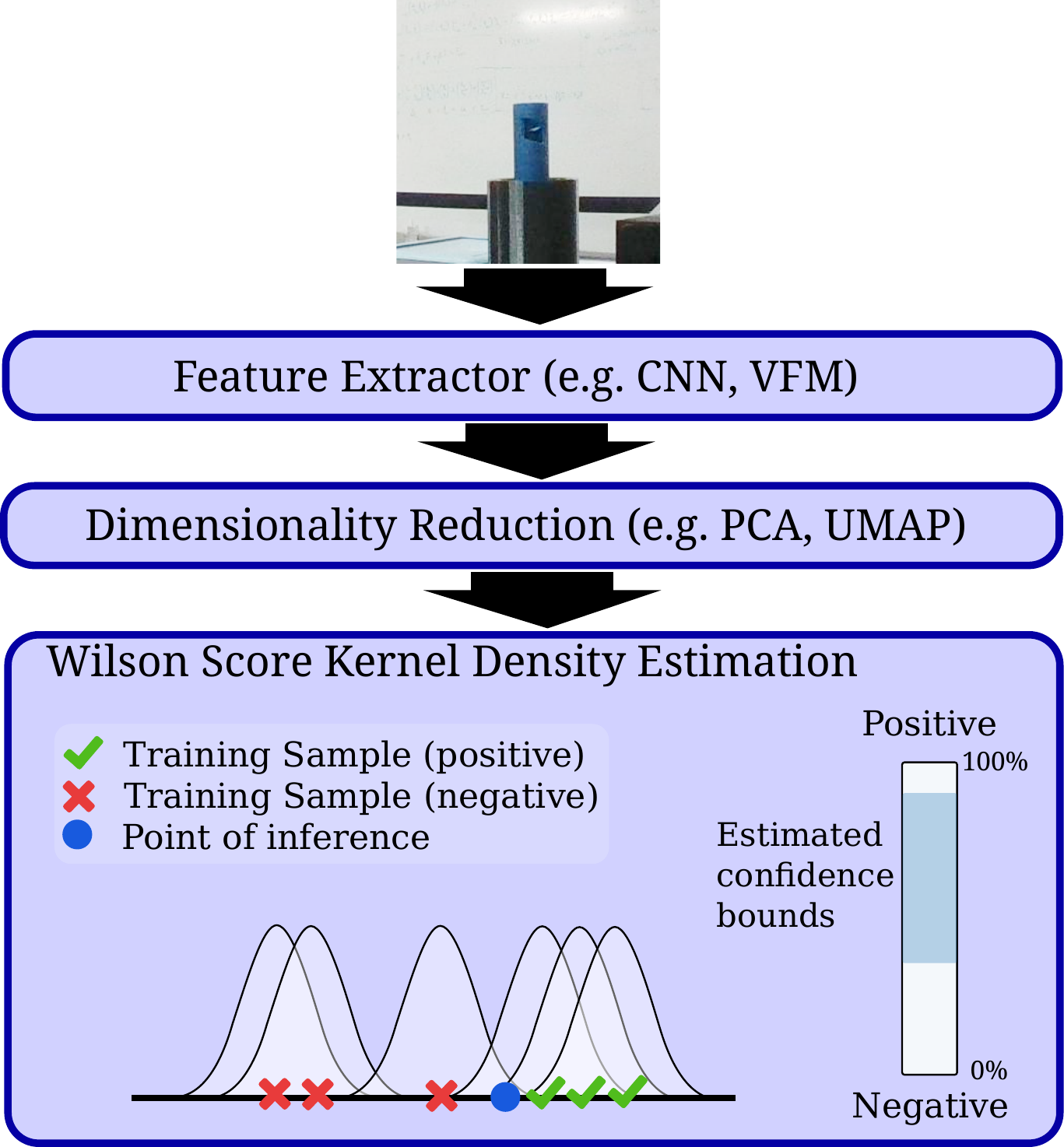}
    \caption{Overview of the Wilson Score Kernel Density Classifier when used for image classification. The image is fed to a feature extractor, e.g. a vision foundation model, and mapped to a lower-dimensional space. Confidence bounds for the class probability are computed from a set of labeled training samples using the Wilson Score Kernel Density Estimator.
    }
    \label{fig:front_page}
\end{figure}

As WS-KDE is a multivariate function estimator, our classification method has wide applicability. Potential uses include classification of force-torque measurements, calibration of confidence scores from trained deep neural networks, and classification of features extracted using foundation models, providing said models with a measure of uncertainty (see Fig.~\ref{fig:front_page}).

The proposed Wilson Score Kernel Density Classifier is essentially a method that, based on a supervised dataset, maps a feature vector to upper and lower confidence bounds on the probability of the predicted class. The method thus allows for:

\begin{itemize}
    \item Binary classification of multivariate signals with associated confidence bounds
    \item Confidence score calibration of binary classifiers with associated confidence bounds
    
\end{itemize}

The paper is structured as follows: Sec. \ref{sec:relatedWork} presents related work. This is followed by Sec. \ref{sec:method}, which presents both a short summary of the WS-KDE function estimator and our method for applying WS-KDE to classification. Sec. \ref{sec:experiments} presents the evaluation, which compares the proposed WS-KDC to Gaussian Process Classification. Finally, Sec. \ref{sec:Conclusion} concludes the paper.

%% file: relatedwork.tex
\section{Related Work}
\label{sec:relatedWork}
The performance of today's image classifiers is outstanding. However, deep learning based classification methods are in many cases not used for critical tasks due to the lack of trust in the models' confidence estimates. The following provides an overview of methods that seek to overcome this by providing reliable uncertainty estimates.

Most probabilistic classifiers are trained to provide a confidence score of their prediction, which can be interpreted as a probability. Confidence calibration is the task of mapping predicted class confidence to the actual probability associated with the class. Standard confidence classification techniques include Platt scaling, temperature scaling, and isotonic regression~\cite{guo2017calibration}. 

In principle, a well-calibrated confidence score is enough to make reliable classifications. However, in practice, there is uncertainty in the calibration process, which must also be quantified. The uncertainty, quantified as confidence bounds on the class probability, can be used to ensure a desired system performance if the classifier is allowed to abstain from making a decision if the associated confidence of the prediction is too low. This is referred to as selective classification~\cite{traub2024overcoming}.

The Venn-Abers predicter~\cite{vovk2012venn} is a wrapper for binary classifiers which provides guaranteed bounds on the class probability. This guarantee depends on the definition of a Venn taxonomy, which associates the inferred point with other points in the calibration set. However, the method does not explicitly account for uncertainty in the sampling of the calibration set.

There are several Bayesian frameworks for estimating uncertainties in classification. A popular one is Gaussian Process Classification, which models the training set using a Gaussian Process. Since the distributions modeled by the Gaussian process are Gaussian, they map the output to a probability distribution using a transfer function~\cite{rasmussen2006gaussian}. Gaussian processes are both theoretically elegant and can, in practice, provide accurate uncertainty estimates if tuned correctly. However, they are computationally expensive to optimize, and thus, practical use of the method for large datasets requires approximate methods. It is notable that Gaussian Process Classification has been used with good results as a classification head of foundational models~\cite{lu2024uncertainty}. 

Bayesian neural networks (BNN) are stochastic networks that are optimized through the application of Bayes' Theorem. The majority of BNNs treat the weights in the network as random variables with an associated joint distribution~\cite{jospin2022hands}. The estimation of the weight distribution is in practice intractable, so various approximations exist, including deep ensembles~\cite{lakshminarayanan2017simple}, Monte Carlo Dropout~\cite{gal2016dropout}, and Variational inference~\cite{blundell2015weight}. The performance of the above networks is generally unstable, with the best method varying from experiment to experiment. Deep ensembles are generally the most consistent but also computationally expensive.

Conformal prediction is another approach to providing guarantees on the reliability of a classification system~\cite{angelopoulos2023conformal}. The idea is to allow the classifier to estimate a set of classes if confidence in a single class is too low, and through this, provide statistical guarantees on the marginal classification coverage. In the context of selective binary classification, a class set of size two corresponds to abstaining from making a decision. While the statistical guarantee is promising, it is important to note that the guarantee is on the marginal coverage, and not on a per-instance basis.

% Gaussian processes provide bounds on the uncertainty estimates, but as a Baysian method is highly reliant on the priors, and the resultant bounds are technically credible intervals, e.g. they express the belief of the outcome, not a rigid confidence score.

Our proposed method is based on the Wilson Score Kernel Density Estimator~\cite{sorensen2020wilson, iversen2025global}, which is a frequentist method for estimating confidence bounds in Binomial experiments that are conditioned on a set of control parameters. The WS-KDE has the desirable property that it only has a single tunable hyperparameter: the bandwidth of a Gaussian kernel. The rest follows from statistical analysis. This means that the WS-KDE provides statistically sound confidence bounds under the assumption that the lengthscale of the kernel is chosen with respect to the smoothness of the feature space. To the best of our knowledge, our proposed method is the first to use Wilson Score Kernel Density Estimation in the context of classification.

%% file: method.tex
\section{Method}
\label{sec:method}
The motivation for estimating confidence bounds on binary classification is to ensure a desired classification performance for critical operations, especially in the context of deep learning based classification of images. However, as Wilson Score Kernel Density Classification is a general classification method with wide applicability, this section is dedicated to providing an intuitive understanding of WS-KDE and framing it in the context of selective classification. A discussion on how to apply our method for image classification is provided in Sec.~\ref{sec:experiments}.

\subsection{Selective binary classification}
In supervised binary classification, a classifier is given a set of training instances $(x_i, y_i)$, $i = 1,\dots, N$, where $x_i\in\mathbb{R}^n$ is a signal and $y_i$ is the target label $y_i \in \left\{ 0,1\right\}$. Here, it has been assumed that the signal is either an N-dimensional vector or that such a vector has been extracted using some form of feature extractor. At inference time, the classifier assigns a label, $\hat{y}\in \{0,1\}$, to a hitherto unseen signal $x_*$. Selective classification, which allows the system to abstain from making a decision if the confidence in the assigned label is too low, is similar to regular classification, except the output label has three possible classes $\hat{y}\in \{0,1, \text{unknown}\}$.

For probabilistic classifiers, classification is often formulated as a regression problem, where the function to be regressed is the conditional probability $S(x)$ of the positive class:

\begin{align}
    S(x) = \conprob{y=1}{x}
\end{align}

% Assuming that our function regressor is calibrated, this formulation provides a straightforward way to set the classification threshold based on the desired minimum success rate $\tau$ of our system:

% \begin{align}
% \hat{y} =
% \begin{cases}
% 1 & \text{if } \hat{S}(x) \geq \tau \\
% 0 & \text{otherwise}
% \end{cases}
% \end{align}

% This is the standard method of classification when using deep probabilistic classifiers.

% Unfortunately, deep learning based classifiers tend to be too optimistic in their assessment of classification confidence. This means that the use of classification 'confidence' becomes a poor metric for assessing system performance. Confidence calibration seeks to mitigate this problem by learning a function $f(\hat{S}(x))$ which maps the estimated confidence to the actual expected success rate, e.i.

% \begin{align}
%     f(\hat{S}(x)) = \mathbb{E}\left[\conprob{y=1}{\hat{S}(x)}\right]
% \end{align}

% Probabilistic classifiers as discussed above provide an estimate of expected performance, but these estimates also have an associated uncertainty. 

There will always be some inherent uncertainty in the regression of $S(x)$ , which in critical operations must be taken into account. Typically, a minimum success rate, $\tau$, is required for the system. When assessing if the classifier's predictions are certain enough to meet this requirement, the assessment must be made for a confidence level of $\alpha$ (e.g. 95\%). If the confidence bounds are defined as $p_\alpha(x) \pm \sigma_\alpha(x)$ the selective classification rule can be expressed as\footnote{ The Wilson Score approximation for computing confidence bounds of a Binomial experiment expresses the bounds using the center of the bounds, $p$, and symmetric bounds around the center $\pm\sigma$, which is why we use this notation here. Note that the center of the bounds, $p(x)$, is equal to the expectation value of the S(x) if and only if the confidence bounds are symmetric around the expectation value, which they generally are not in binomial experiments.}:

\begin{align}
\hat{y} =
\begin{cases}
1 & \text{if } p_\alpha(x)-\sigma_\alpha(x) > \tau \\
0 & \text{if } p_\alpha(x)+\sigma_\alpha(x) < \tau \\
\text{unknown} & \text{otherwise}
\end{cases}
\label{eq:selectiveTau}
\end{align}

A reliable classification system can thus be ensured by employing the above selective classification rule, given that the provided uncertainty bounds are accurate. We argue that WS-KDE can be used to infer such bounds under the assumption that the feature space is smooth and the bandwidth of the smoothing kernel has been chosen accordingly.

\subsection{Wilson Score Kernel Density Estimation}
WS-KDE was introduced in the context of Bayesian Optimization to optimize industrial vibratory feeder traps. While the trap parameters could be controlled, the outcomes of individual experiments were essentially random Bernoulli trials due to uncontrollable perturbations in real-world experiments. Global optimization of the control parameters thus required optimization of the $p$-value, which was a function of the control parameters, $x$. WS-KDE was used as a function estimator in Bayesian optimization to estimate upper and lower confidence bounds on the function $p(x)$.

\begin{figure}
    \centering
    \begin{subfigure}[t]{\linewidth}
    \centering
    \includegraphics[width=\linewidth]{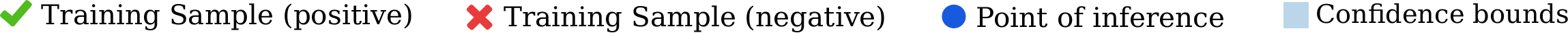}
    \vspace{1pt}
    \end{subfigure}
    \begin{subfigure}[t]{0.48\linewidth}
    \centering
    \includegraphics[width=\linewidth]{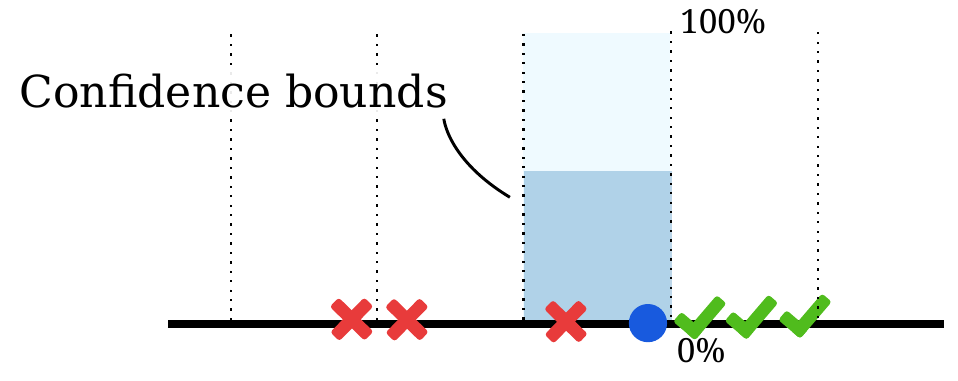}
    \caption{Binning + Wilson Score}
    \end{subfigure}
    \begin{subfigure}[t]{0.48\linewidth}
        \centering
        \includegraphics[width=\linewidth]{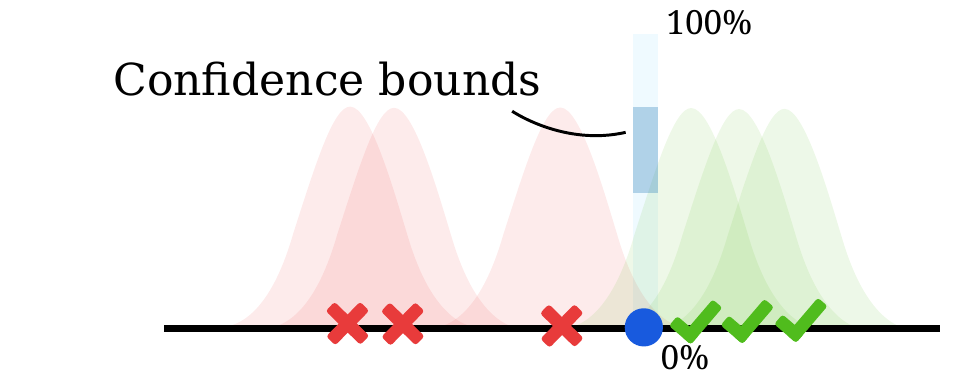}
        \caption{Wilson Score KDE}
    \end{subfigure}
    \caption{Intuitive illustration of WS-KDE. (a) The Wilson Score method could in principle be used for estimating confidence bounds by binning the feature space and treating the samples in each bin as a part of separate Binomial tests. (b) WS-KDE provides a more elegant solution by combining the Wilson Score method with kernel smoothing.}
    \label{fig:wskde_explanation}
\end{figure}

While WS-KDE is arguably quite intuitive, it is perhaps best explained by first looking at a simpler, alternative method. For a visualization, see Fig.~\ref{fig:wskde_explanation}a. First, we divide the feature space into a regular grid of bins. Assuming that the grid is fine enough, we can consider all points that fall into the same bin as being Bernoulli trials from a single Binomial experiment with a single $p$-value. With the data from this Binomial experiment, we can compute the upper and lower bounds on the $p$-value using the Wilson Score method~\cite{wilson1927probable}. In this context, the Wilson Score approximation is preferable to the normal approximation, since it excels at providing proper confidence bounds even for a low number of observations. When we observe a new random sample, we can simply assign the confidence bounds associated with the bin in which it falls.

WS-KDE is an extension of this idea. As illustrated in Fig.~\ref{fig:wskde_explanation}b, WS-KDE uses kernel smoothing to do weighted aggregation of the neighboring points and fuses this with the Wilson Score method.

The only assumptions of the WS-KDE are that the aggregated samples are independent and identically distributed, which means that the function $p(x)$ must be approximately constant over the lengthscale defined by the bandwidth matrix of the Gaussian kernel. This creates a trade-off, where a narrow kernel will lead to less smoothing and consequently larger confidence bounds, while a large kernel will aggregate more samples, leading to faster convergence of the confidence bounds, at the risk of over-smoothing, which leads to invalid confidence bounds. It is still an open question how to choose an optimal kernel bandwidth.

\subsection{WS-KDE for binary classification}

To leverage the properties of WS-KDE, we propose to formulate the classification problem as a random process where the 'control parameter' is the signal being classified and the outcome of the classification is the Bernoulli trial. In the following, we assume that a robotic system performs an action, e.g. an insertion for assembly, which can either fail or succeed. Furthermore, it is assumed that the system is equipped with a sensor (e.g. a vision sensor) which records the process and, through some form of feature extraction (e.g. a neural network) outputs an N-dimensional feature vector $x$. Every time the robot performs the action, it samples a single sample from an unknown distribution $(x_*,y_*) \sim \prob{x,y}$. If the distribution is written as

\begin{align}
    \prob{x,y} = \prob{x}\conprob{y}{x}
\end{align}

it is clear that the sampling process can be expressed as first sampling the feature, followed by sampling of the label conditioned on the feature. While the label for a given instance is not random, it must be noted that the feature in most practical cases will be the output of a feature extractor, which cannot be relied upon to perfectly map sensor input to unique signals with associated known success or failure. Thus, for a given signal, there will be associated aleatoric uncertainty, which can be interpreted as a Bernoulli trial with a \textit{p}-value that will depend on the data distribution, e.i. a combination of the underlying distribution of successes and failures and how well the feature extractor separates these cases. Using WS-KDE, the confidence bounds of the $p(x)$ function can be estimated and interpreted as confidence bounds.

An important observation from the above discussion is that the accuracy of the confidence bounds depends on the choice of bandwidth matrix and the correctness of the i.i.d. assumption. Most importantly, the accuracy of the confidence bounds is not reliant on the performance of the feature extractor, as poor performance will impact the $p$-value, not the accuracy of the confidence bounds. This means that WS-KDE can be used as a classification head for any feature extractor. This is essential when the feature extractor is trained on out-of-domain data, such as synthetic images, or when using a foundation model as the feature extractor.

%% file: experiments.tex
\section{Experiments}
\label{sec:experiments}

\begin{figure}
    \centering
    \begin{subfigure}[t]{\linewidth}
    \centering
    \includegraphics[width=\linewidth]{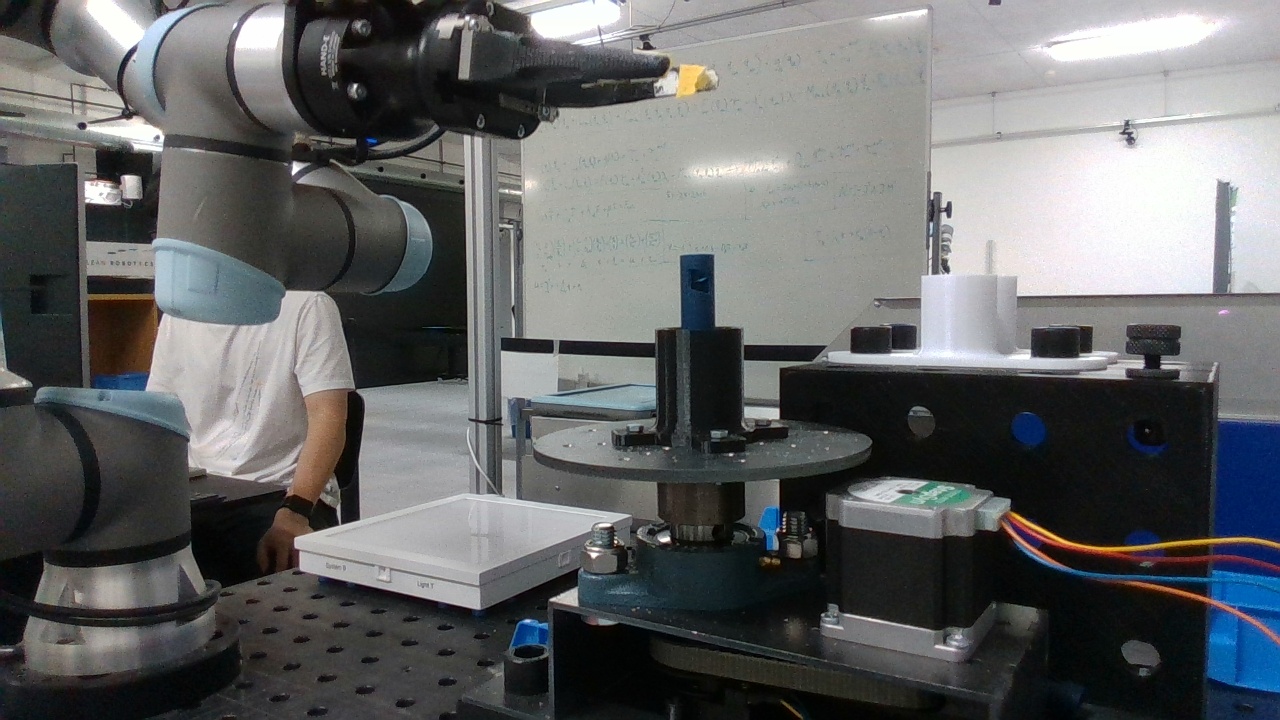}
    \caption{Setup used for automated assembly\\\ }
    \end{subfigure}
    \begin{subfigure}[t]{0.49\linewidth}
        \centering
        \includegraphics[width=0.47\linewidth]{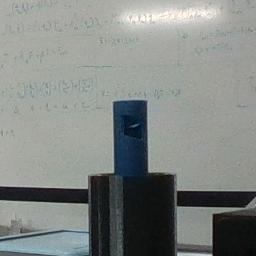}
        \includegraphics[width=0.47\linewidth]{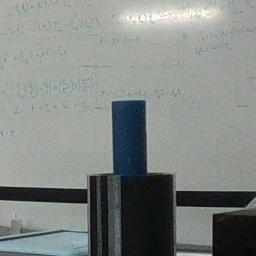}
        \hfill
        \caption{Successful insertion}
    \end{subfigure}
    \hfill
    \begin{subfigure}[t]{0.49\linewidth}
        \centering
        \includegraphics[width=0.47\linewidth]{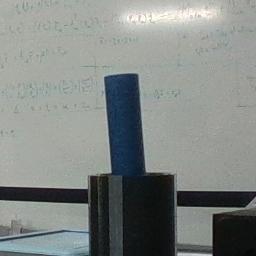}
        \includegraphics[width=0.47\linewidth]{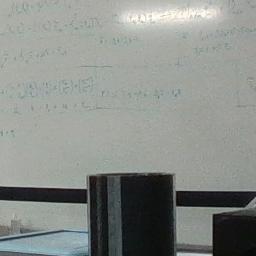}
        \caption{Failed insertion}
    \end{subfigure}
    \caption{As part of an automated assembly, the robot inserts a part into a fixture. A vision-based classifier is used to infer in the insertion was successful, before the assembly process can continue.}
    \label{fig:assembly}
\end{figure}
The evaluation compares the proposed Wilson Score Kernel Density Classifier (WS-KDC) with a Gaussian Process Classifier (GPC) in the context of selective classification. First, the evaluation metrics will be introduced, followed by a description of the evaluation procedure and the dataset used for the evaluation. The results of the evaluation are presented at the end of the section.

\subsection{Evaluation metrics}
One of the challenges of evaluating uncertainty estimates is that the test split only contains the true labels, not the true uncertainty estimates. For probabilistic classifiers that output single class confidences, the calibration of said classifiers can be quantified using e.g. Biar-score, Expected calibration error, or negative log-likelihood~\cite{guo2017calibration}. If the classifier instead predicts a distribution over confidence scores, the expectation value of the given metric can be computed instead. However, the output of WS-KDC deviates from standard classification methods in that it only provides upper and lower confidence bounds on the class confidence without providing an expectation value or distribution over the confidences. An alternative would be to use the midpoint between the bounds as a proxy for expectation value, however, this would implicitly approximate the probability distribution as uniform, which would be a poor approximation to use in an evaluation.

% for ex This means the above metrics that standard metrics for evaluating unceratinty estimates, such as negative log-likelihood (NLL), classifier calibration such as Biar-score~\cite{briar}, Expected calibration error (ECE)~\cite{briar}, and negative log-loss (NLL)~\cite{briar} require an estimate of the success rate, and does not have a corresponding method when only the bounds are provided. One might be tempted to simply use the midpoint of the bounds, or as in \cite{maybeWennAber} use the success rate that minimizes regret. However, this approach would heavily and unfairly penalize the WS-KDE classifier for providing wide confidence bounds in the neighbourhood of isolated samples in the feature space. If e.g. a single success was observed in a given region of feature space, one would expect that a new sample in the same neighbourhood would also be a success, given no other information. However, the Wilson Score prediction would rightly predict wide confidence bounds, since from a statistical standpoint, there is too little data to make a certain statement about the expected class.

\begin{figure*}
    \centering
    \begin{subfigure}[t]{0.24\linewidth}
        \centering
        \includegraphics[width=\linewidth]{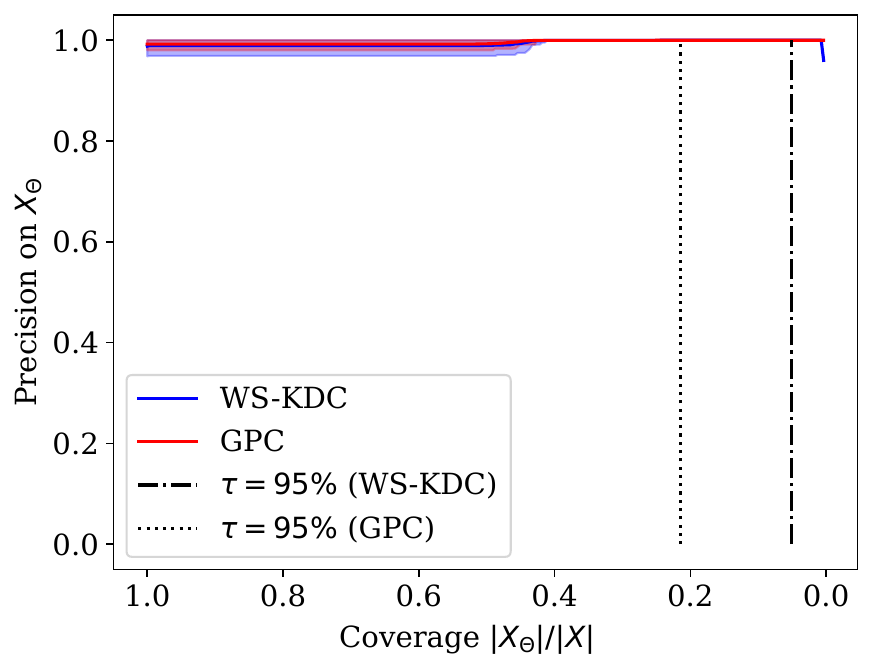}\\
        \includegraphics[width=\linewidth]{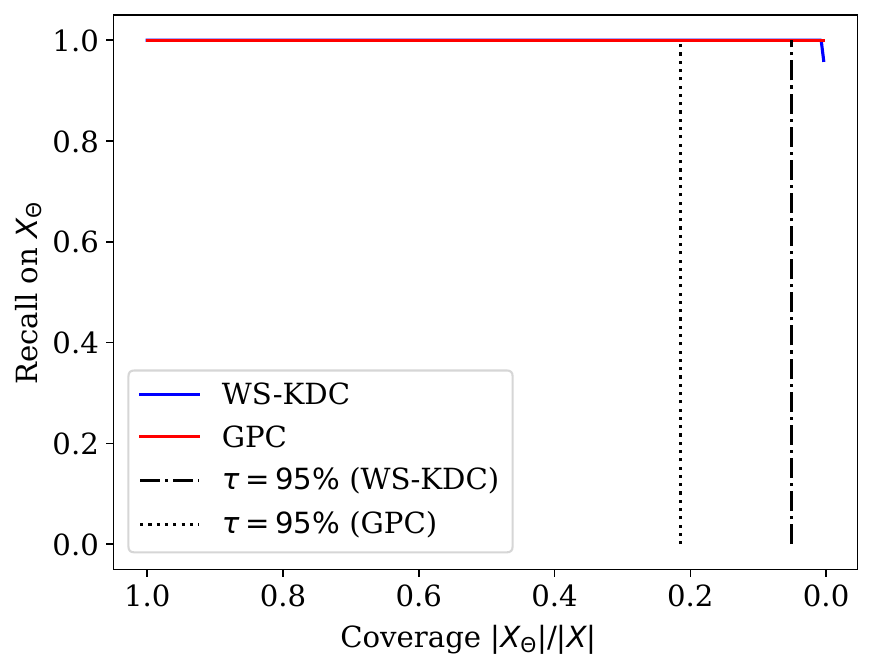}
        \caption{Bank}
    \end{subfigure}
    \begin{subfigure}[t]{0.24\linewidth}
        \centering
        \includegraphics[width=\linewidth]{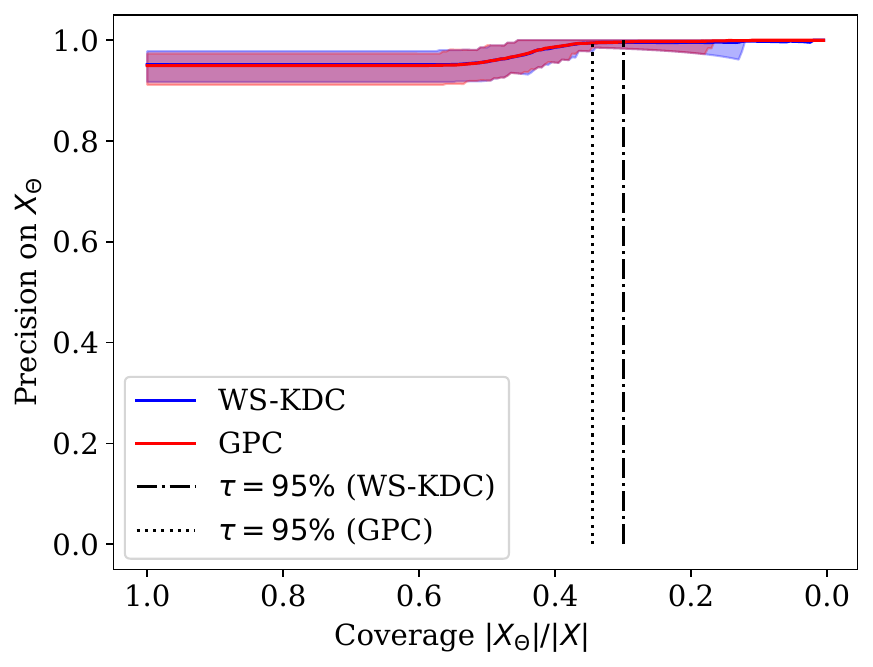}\\
        \includegraphics[width=\linewidth]{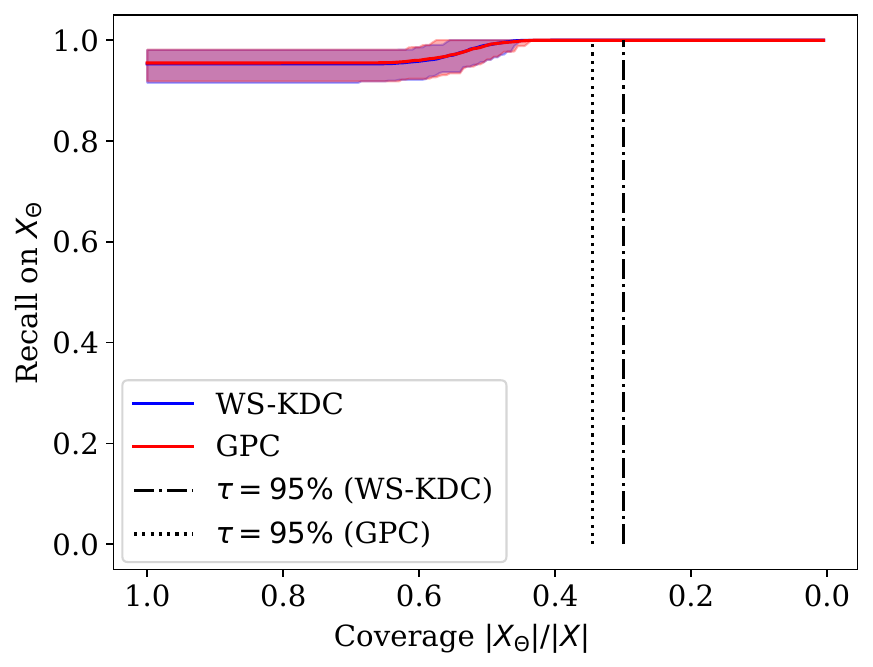}
        \caption{Cats \& Dogs (1k)}
    \end{subfigure}
    \begin{subfigure}[t]{0.24\linewidth}
        \centering
        \includegraphics[width=\linewidth]{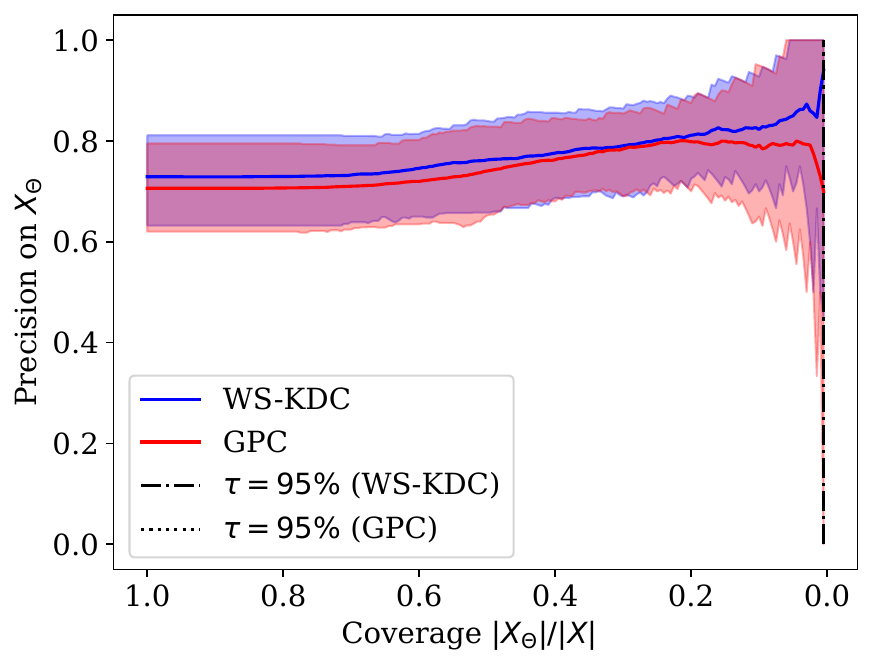}\\
        \includegraphics[width=\linewidth]{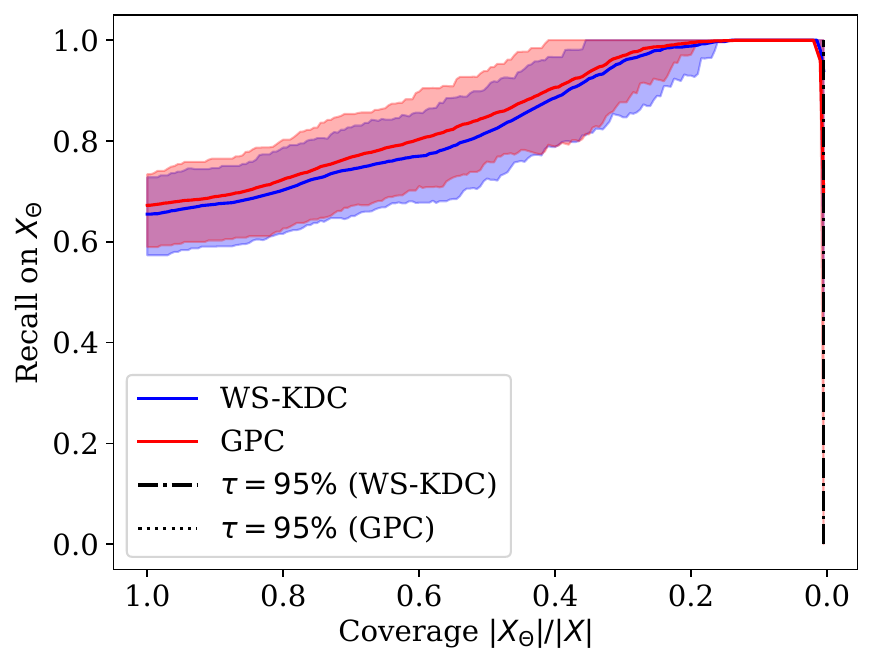}
        \caption{chestMNIST (1k)}
    \end{subfigure}
    \begin{subfigure}[t]{0.24\linewidth}
        \centering
        \includegraphics[width=\linewidth]{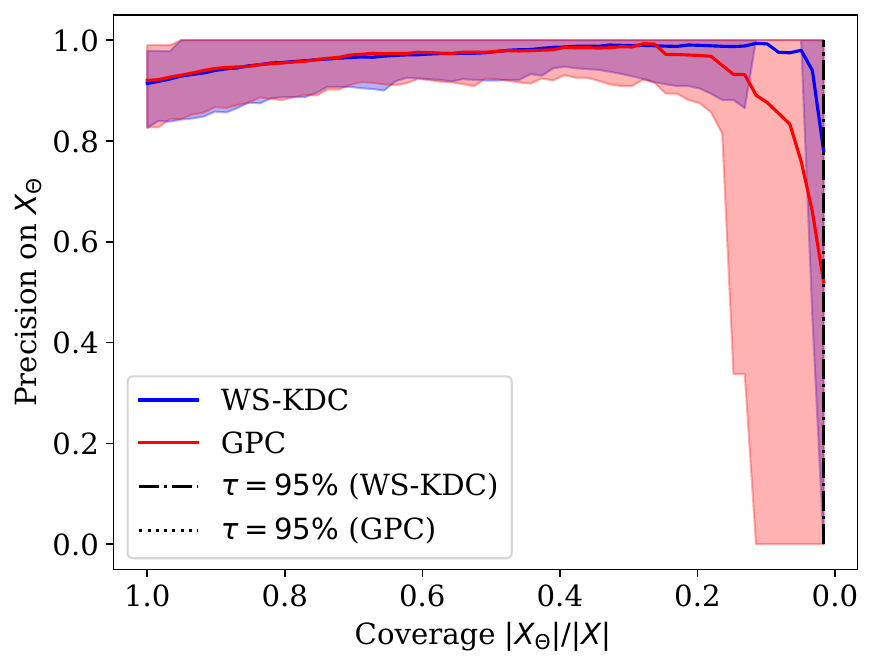}
        \includegraphics[width=\linewidth]{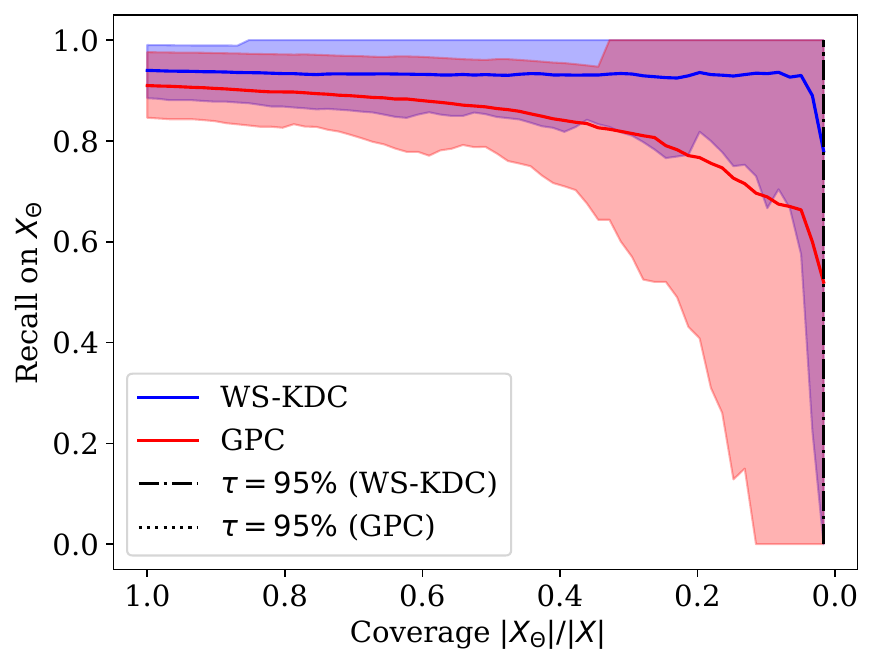}
        \caption{Assembly}
    \end{subfigure}
    \caption{Precision/Recall reject plots for four of the experiments. In plots b and c ResNet18 is used as the feature extractor, while plot d uses Dinov3. In b-d dimensionality reduction is done with UMAP. The lines indicate the mean, and the shaded area indicates the 5\% and 95\% quantiles over 50 repeated experiments. The vertical lines indicate at which coverage the corresponding lower confidence bound is at $\tau = 95\%$. The precision to the right of this line must be above $\tau$ for the classifier to be reliable.}
    \label{fig:prc_rrc}
\end{figure*}

For this reason, the WS-KDC is evaluated in the context of selective classification. Since there is currently no agreed-upon standard metric for selective classification~\cite{traub2024overcoming}, we follow the suggestion in~\cite{fischer2023precision} and plot the prediction reject curve (PRC) and recall reject curve (RRC), which plot the selective prediction and recall against the selective coverage (1-reject rate). Selective precision and recall are defined as the precision and recall computed for the subset of the datapoints $X_\Theta\in X$ that has been selected, i.e. are confident enough not to be rejected. In practice, this is computed by classifying all the samples in the test set and assigning a measure of classification confidence to each sample. The coverage corresponding to rejection of the $i$ least confident classifications is thus $c_i = (N-i)/N$ for $i\in{0,\dots,N-1}$. Since the selective precision and recall can be computed for each value of the coverage, the area under the curves can be computed numerically using a Riemann sum:

\begin{align}
AUPRC &\approx \frac{1}{N}\sum_{i=0}^{N-1}\frac{\text{TP}_i}{\text{TP}_i + \text{FP}_i} \\
AURRC &\approx \frac{1}{N}\sum_{i=0}^{N-1}\frac{\text{TP}_i}{\text{TP}_i + \text{FN}_i}
\end{align}

where $\text{TP}_i$, $\text{FP}_i$, and $\text{FN}_i$ are respectively the number of true positives, false positives, and false negatives when the $i$ least confident classifications are rejected.

We choose the predicted class based on the midpoint between the upper and lower bounds and assign the confidence as follows:
\begin{align}
\text{confidence} =
\begin{cases}
p_\alpha(x)-\sigma_\alpha(x) &\text{if}\ p_\alpha(x) >= 0.5\\
p_\alpha(x)+\sigma_\alpha(x) & \text{otherwise}
\end{cases}
\end{align}

This choice of confidence corresponds with the selective classifier in Eq.~\ref{eq:selectiveTau}. Since the connection between coverage and desired success rate $\tau=95\%$ is not immediately apparent from the PRC and RRC curve, we add an indicator of the coverage corresponding to a lower confidence bound of $95\%$.

\subsection{Evaluation pipeline}
As mentioned in Sec. \ref{sec:relatedWork}, Gaussian Process Classification (GPC) has been used with good results as the classification head of foundational models~\cite{lu2024uncertainty}, which is similar to our suggested use of WS-KDC. The method in~\cite{lu2024uncertainty} uses a foundation model for extracting features, a trained linear projector for dimensionality reduction, and a Gaussian Process Classifier with an approximate Dirichlet likelihood for classification. As the implementation in~\cite{lu2024uncertainty} is not readily available, we implement a similar pipeline consisting of a feature extractor, a dimensionality reduction method, and a classification head. For the classification head, we compare a Gaussian Process Classifier with a Bernoulli likelihood~\cite{scikit-learn} to our proposed Wilson Score Kernel Density Classifier. The two classifiers are evaluated on four different datasets with varying choices of feature extractor and dimensionality reduction method. An overview of these is provided below.

Both the WS-KDC and the GPC require tuning of hyperparameters. We choose an isotropic Gaussian kernel for both methods, which means that both methods require the length scale of the kernel to be optimized, with the GPC also requiring optimization of the kernel variance. For the WS-KDC, the lengthscale is optimized with 10-fold cross-validation of the training set, using the negative log likelihood of the held out samples as loss. The likelihood is computed using a Bernoulli likelihood, with the $q$ value chosen as the midpoint between the estimated upper and lower bounds. Using the midpoint strikes a balance between smoothing and accuracy, since the best-case confidence bound can be trivially optimized by choosing a narrow kernel, and the worst-case bound will prioritize smoothing. As WS-KDE only requires optimization of the length scale, we use a 20-step logarithmic linesearch from $10^{-2}d$ to $10^{-1}d$, where $d$ is the mean distance between all point pairs in the training set. The GPC is optimized using the L-BFGS-B optimizer in~\cite{scikit-learn} with variance initialized to 1, and length scale initialized to $10^{-1}d$. For computing the confidence bounds of the GPC, we sample the posterior 100 times and compute the $\alpha/2$ and $1-\alpha/2$ percentiles as the lower and upper bounds, respectively.

The evaluation covers four different datasets, with various methods of feature extraction and dimensionality reduction. The feature extractor is dataset dependent, while the dimensionality reduction methods are data agnostic. The following section will therefore present the four datasets and associated feature extractors together, followed by an overview of the dimensionality reduction methods.

\begin{table*}[]
    \centering
    \begin{tabular}{|lll|cc|cc|}
    \hline
     &  & & AUPRC $\uparrow$ & AURRC $\uparrow$ & $t_\text{optim}\ (s)\downarrow$ & $t_\text{infer}\ (s)\downarrow$\\
     Dataset& Dim. reduction & Method &  mean [std] & mean [std] & mean [std] & mean [std] \\
    \hline
     Bank~\cite{banknote_authentication_267} & None & GPC &  0.996 [0.004] & 1.0 [0.0] & 12.1 [0.7] & 0.04 [0.01]\\
     &  & WS-KDC   &   0.994 [0.006] & 0.9999 [0.0007]& \textbf{0.23 [0.03]} & 0.05 [0.02]\\
     \hline
     Cats \& Dogs~\cite{cats_and_dogs} (1k) & ResNet18 + pca& GPC & 0.97 [0.01] &0.98 [0.01] & 4.6 [0.4] & 0.03 [0.01]\\
                      & &WS-KDC & 0.97 [0.01] & 0.98 [0.01]& \textbf{0.22 [0.01]}& 0.04 [0.03]\\
    \cline{2-7}
    & ResNet18 + umap& GPC & 0.97 [0.01]& 0.99 [0.01]& 4.5 [0.4] & 0.03 [0.01]\\
           & &WS-KDC &0.97 [0.01] & 0.98 [0.01]& \textbf{0.14 [0.01]}& 0.06 [0.03]\\
    \cline{2-7}
    & ResNet18 + fc& GPC & 0.97 [0.01] & 0.978 [0.008] & 4.4 [0.6]& 0.03 [0.01]\\
           & &WS-KDC  & 0.97 [0.01]& 0.98 [0.01]& \textbf{0.15 [0.02]} & 0.057 [0.03]\\
    \hline
     Cats \& Dogs~\cite{cats_and_dogs} (5k) & ResNet18 + umap &WS-KDC & 0.971 [0.007]& 0.983 [0.005] & \textbf{2.3 [0.2]}& 0.047 [0.003]\\
    \hline
    Chestmnist~\cite{medmnistv2} (1k) & ResNet18 + umap& GPC &0.75 [0.05] & 0.85 [0.04] & 4.0 [0.5] & 0.03 [0.01]\\
                      & &WS-KDC & 0.77 [0.05] & 0.84 [0.04] & \textbf{0.14 [0.01]}& 0.04 [0.03] \\
    \cline{2-7}
    & ResNet18 + fc& GPC & 0.74 [0.04] & 0.88 [0.03]& 4.4 [0.5]& 0.03 [0.01]\\
           & &WS-KDC & 0.74 [0.05]& 0.84 [0.04]& \textbf{0.14 [0.01]}& 0.04 [0.03]\\
    \hline
    Chestmnist~\cite{medmnistv2} (22k) & ResNet18 + umap & WS-KDC & 0.74 [0.01] & 0.863 [0.007] & \textbf{79.8 [1.7]} & 1.02 [0.05]\\
    \hline
    Assembly Inspection & Dinov3 + umap& GPC & 0.94 [0.05] & 0.83 [0.1] & 0.30 [0.07] & 0.013 [0.005]\\
                      & &WS-KDC & 0.97 [0.03] &  0.93 [0.05] & \textbf{0.077 [0.004]}& 0.04 [0.04]\\
     & Dinov3 + logReg & GPC & 0.99 [0.01] & 0.99 [0.01] & 0.34 [0.09]& 0.016 [0.005]\\
                      & &WS-KDC & 0.98 [0.02] & 0.98 [0.02]& \textbf{0.14 [0.01]}& 0.004 [0.005]\\
    \hline
    \end{tabular}
    \caption{Summary of experimental results for WS-KDC and GPC. The experiments cover four datasets and various feature extractions and dimensionality reduction methods. The data shows that the selective classification performance expressed as the area under the precision/recall reject curve (AUPRC/AURRC) was similar for the two methods. The only considerable difference between the two methods is the optimization time, where WS-KDE is significantly faster.}
    \label{tab:results}
\end{table*}

\subsubsection{Dataset: Banknote Authentication~\cite{banknote_authentication_267}}
The data in the 'Banknote Authentication' dataset are features extracted from wavelet-transformed images of authentic and forged bank notes. There are 1372 samples with 4 features each. Since a wavelet transform-based feature extractor was already used in the construction of the dataset, no further processing was required.

\subsubsection{Dataset: Cats and Dogs~\cite{cats_and_dogs}}
The data in 'Cats and Dogs' consists of 25000 images, each labeled as either cat or dog. We used 20000 of the images to train a Resnet18~\cite{resnet} for the task for binary classification\footnote{Backbone: ResNet18 pretrained on ImageNet, classification head: 1 fully connected, Optimizer: Adam, loss: binary cross entropy, learning rate: $10^{-3}$, batch size: 32, epochs: 2, transforms: ImageNet normalization, train-validation split: 0.8/0.2}. The trained network was then used as the feature extractor on the remaining 5000 images either by extracting the 512-dimensional flattened features of the last convolutional layer (ResNet18), or by using the entire network to extract class probability and treating it as a 1D feature (ResNet18 + fc). We refer to the full set of extracted features as 5k, and a subset of 1000 images randomly sampled from the dataset as 1k.

\subsubsection{Dataset: ChestMNIST~\cite{medmnistv2}}
The ChestMNIST dataset consists of 112,120 frontal-view X-ray images of patients. Each image is provided with a binary label for each of 14 possible diseases. We use the dataset for binary classification by labeling the patient as healthy if and only if no disease was present. We trained a Resnet18 for feature extraction using the designated 78,468 training and 11,219 validation images\footnote{Backbone: ResNet18 pretrained on ImageNet, classification head: 1 fully connected, optimizer: Adam, loss: binary cross entropy, learning rate: $10^{-4}$, batch size: 64, epochs: 5 + early stopping, transforms: \{mnist normalization, 3-fold stacking, horizontal flip, random rotation ($10^\circ)$\}, train-validation split: as designated by dataset.}. The trained network was then used to extract features from the last 22,433 images. Similar to the cats and dogs dataset, we use both the extracted features (resnet18) and the estimated class probability (resnet+fc) as feature extractors. We refer to the full set of extracted features as 22k, and a subset of 1000 images randomly sampled from the dataset as 1k.

\subsubsection{Dataset: Assembly inspection}
The last dataset consists of images taken in our lab during a robotic assembly. For each action, the robot inserts a part into a fixture, and an image of the insertion is recorded using an RGB camera. If the part is not inserted correctly, subsequent actions will fail, so binary classification is used to assess if the insertion was successful. To evaluate the use of the classifiers in the context of a foundation model, we employ Dinov3 ConvNeXT-tiny~\cite{simeoni2025dinov3} as feature extractors. This results in a dataset consisting of 303 samples with 768-dimensional features. Example images from the dataset are shown in Fig. \ref{fig:assembly}

\subsubsection{Dimensionality reduction methods}
We evaluate three different dimensionality reduction techniques. The first is principal component analysis (PCA), which does a linear mapping of the features into a new coordinate frame such that the mapped features correspond to the largest variations in the dataset. The output features are the $D$ features corresponding to the largest principal components.

The second dimensionality reduction technique is umap~\cite{umap}, which is a non-linear dimensionality reduction technique that aims to reduce dimensionality in a way that preserves the local structure of the high-dimensional space. This makes it theoretically well-suited for use with Gaussian kernels as the lengthscale tends to be low enough that only the local neighbourhood distances are relevant. For the umap paraemters we use $\text{n\_neighbours}=15$ and $\text{min\_dist} = 0.1$.

The final dimensionality reduction technique is that of using a linear classifier. The use of the linear layers in the ResNet18 networks has already been mentioned. For the Dinov3 model, we use logistic regression.

To keep the evaluations consistent, the output dimensionality of both PCA and UMAP is set to $D=3$ for all experiments.

\subsubsection{Estimation of evaluation variance}
The evaluation of WS-KDE and the GPC uses an 80\%/20\% train/test split. In order to investigate the variance in the performance of the methods, we run each experiment 50 times using different random seeds to make the train/test split in each experiment. In this way, we can investigate the variance in performance caused by random sampling without leakage between the test and train set.

\subsubsection{Results}
A subset of the Precision Reject Curves and Recall Reject Curves is shown in Fig.~\ref{fig:prc_rrc}. While the curves are different for each dataset, they show an increase in performance as the reject rate increases, i.e. the coverage decreases. This shows that the use of the estimated confidence bounds to decide for rejection indeed increases performance as expected.

There are a few observations that should be commented on. Both the cats \& dogs and chestMNIST datasets use a trained resnet18 for feature extraction. However, the precision and recall for the cats \& dogs dataset are much higher. This is not surprising, as it is an easier classification problem to solve. However, the vertical $\tau=95\%$ line shows that for the cats and dogs dataset, both classifiers reach a coverage level where confident predictions can be made. However, for the chestMNIST, both classifiers find that none of the images can be classified with the required confidence.

For the Assembly dataset, we observe that Dinov3 is able to extract relatively good features for classification, however, the precision and recall fall far below the desired threshold at low coverage. A qualitative investigation into the cause of this showed that the low amount of data used when optimizing the lengthscale of the kernels led to an unreasonably large amount of smoothing, causing confidence bounds to be too optimistic. This highlights the importance of choosing a proper kernel lengthscale for balancing smoothing and prediction accuracy. Other ways to choose the bandwidth matrix, besides the optimization procedure proposed in this paper, is left for future work.

\begin{figure}
    \centering
    \includegraphics[width=\linewidth]{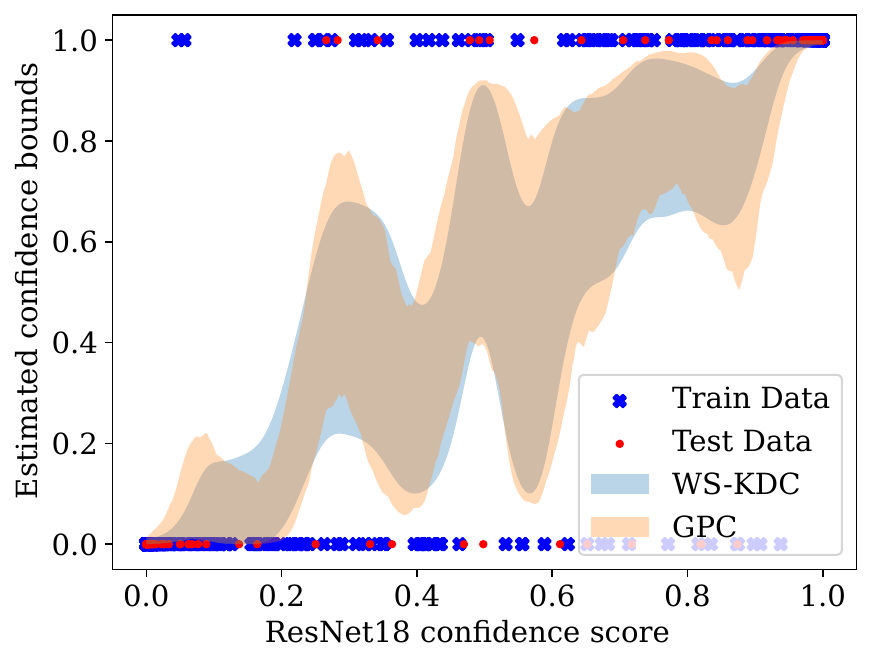}
    \caption{Plot of the mapping from ResNet18 classification confidences to the confidence bounds estimated by the WS-KDC and GPC during a single run of the \textit{Cats \& Dogs ResNet18 + fc (1k)} experiment.}
    \label{fig:stik_proeve}
\end{figure}

Performance-wise, the curves for WS-KDC and GPC are similar, with none of the methods clearly outperforming the other. Our hypothesis is that the two methods generally arrive at approximately the same confidence bounds despite their differences in methodology. As a qualitative test of this hypothesis, we visualized the estimated confidence bounds as a function of the 1D feature extracted from the Cats and Dogs (1k) ResNet18 + fc network (see Fig.~\ref{fig:stik_proeve}). The fact that the bounds predicted by the two methods are relatively similar supports our hypothesis.

Quantitative metrics for all experiments across all datasets are shown in Table~\ref{tab:results}. Similar to the observations from the curves, there is no classification method that outperforms the other in terms of selective classification performance. However, WS-KDC outperforms GPC when it comes to the optimization time of hyperparameters. The difference is large enough that we had to introduce the 1k dataset variants in order to do the evaluation on both methods in a reasonable timespan. The full 5k and 22k datasets are therefore only done for WS-KDE.

\begin{figure}
    \centering
    \includegraphics[width=0.48\linewidth]{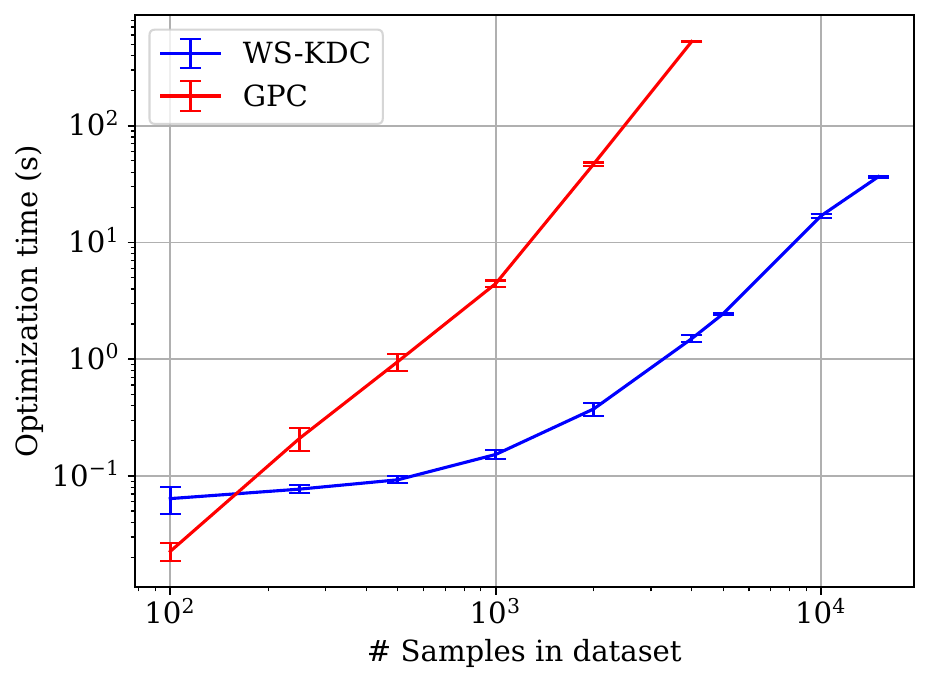}
    \includegraphics[width=0.48\linewidth]{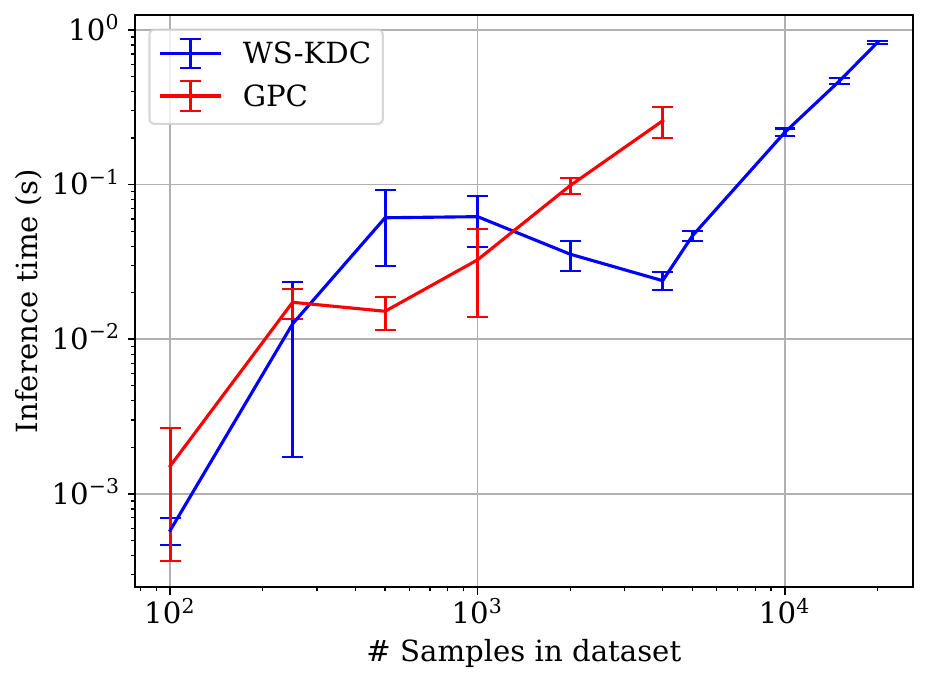}
    \caption{Double logarithmic Plot of the mean of the optimization and inference time of WS-KDC and GPC when applied to various-sized subsets of the \textit{chestMNIST ResNet18 + umap} experiment. Error bars indicate the standard deviation. The plot stops at 4000 for GPC as the mean optimization time was 525 seconds. In comparison, WS-KDC was optimized in 1.5 seconds.}
    \label{fig:timing}
\end{figure}

To compare optimization and inference time between the two classifiers, we ran 10 experiments with different subsets of the chestMINST resnet18 + umap. This is plotted in a double-logarithmic plot in Fig.~\ref{fig:timing}. Note that the plot only goes up to 4000 datapoints for the GPC since it took an average of 525 seconds to optimize. WS-KDC was optimized in 1.5 seconds, making it over two orders of magnitude faster than GPC.

All experiments were done on an AMD Ryzen 7 8840U CPU. We point out that optimization of the lengthscale in WS-KDE was done using linesearch and cross-validation, making optimization of the process trivially parallelizable. Furthermore, while the implementation of WS-KDE used in the experiments~\cite{iversen2025global} did not have a GPU implementation available, it should be noted that the WS-KDE method is highly parallelizable. The theoretical runtime for a GPU-accelerated version of WS-KDC is thus much lower than what we report in these experiments.

%% file: conclusion.tex
\section{Conclusion} 
\label{sec:Conclusion}
In this work, we have proposed a novel method for estimating the confidence bounds in binary classification. The method is based on the Wilson Score Kernel Density Estimator, which combines kernel density smoothing with the Wilson Score method for estimating confidence bounds in binomial experiments. Our method was evaluated on four different datasets in the context of selective classification and compared to Gaussian Process Classification. The results showed similar selective classification performance between the two methods. However, our proposed method was significantly faster, has fewer hyperparameters to tune, and is arguably more intuitive than Gaussian Process Classification.